\documentclass[11pt]{article}
\usepackage[a4paper]{geometry}
\usepackage{latexsym}
\usepackage{authblk}
\usepackage{clic2017}
\usepackage{times}
\usepackage{url}
\usepackage{latexsym}

\usepackage{paralist}
\usepackage{amsmath}
\usepackage{array}
\usepackage{tabularx}
\usepackage{multirow}
\usepackage{graphicx}
\usepackage{amsfonts}
\usepackage{amssymb}
\usepackage{url}
\usepackage{tcolorbox}
\usepackage{enumitem}
\usepackage{graphicx}
\usepackage{multirow, makecell}
\usepackage{lingmacros}
\usepackage{bbm}
\usepackage{booktabs}
\usepackage[T1]{fontenc}
\usepackage[outline]{contour}
\usepackage{xcolor}
\contourlength{0.2pt}
\contournumber{55}%
\newcolumntype{C}{>{\centering\arraybackslash}X}

\usepackage[caption=false]{subfig}

\usepackage[T1]{fontenc}
\usepackage[outline]{contour}
\usepackage{xcolor}
\contourlength{0.2pt}
\contournumber{55}%

\usepackage[utf8]{inputenc}
\usepackage{amsmath}
\usepackage{fourier}
\usepackage{esvect}
\usepackage{scalerel}

\usepackage{times}
\usepackage{helvet}
\usepackage{courier}
\usepackage{graphicx} 
\usepackage{fancyhdr}
\usepackage{etoolbox}
\usepackage[export]{adjustbox}
\usepackage{float}
\usepackage{amsmath}

\title{QMUL-SDS @ DIACR-Ita: Evaluating Unsupervised Diachronic Lexical Semantics Classification in Italian}

\author[1,2]{\textbf{Rabab Alkhalifa}}
\author[1,3]{\textbf{Adam Tsakalidis}}
\author[1]{\textbf{Arkaitz Zubiaga}}
\author[1,3]{\textbf{Maria Liakata}}
\affil[1]{Queen Mary University of London, United Kingdom}
\affil[2]{Imam Abdulrahman bin Faisal University, Saudi Arabia}
\affil[3]{Alan Turing Institute, United Kingdom}

\date{}

\begin{document}
\maketitle
\begin{abstract}
In this paper, we present the results and main findings of our system for the DIACR-Ita 2020 Task. Our system focuses on using variations of training sets and different semantic detection methods. The task involves training, aligning and predicting a word's vector change from two diachronic Italian corpora. We demonstrate that using Temporal Word Embeddings with a Compass C-BOW model is more effective compared to different approaches including Logistic Regression and a Feed Forward Neural Network using accuracy. Our model ranked 3rd with an accuracy of 83.3\%.
  
\end{abstract}
{\let\thefootnote\relax\footnotetext{Copyright \textcopyright\ 2020 for this paper by its authors. Use permitted under Creative Commons License Attribution 4.0 International (CC BY 4.0).}} 

\section{Introduction}
The quantitative analysis of language evolution over time is a new emerging research area within the domain of Natural Language Processing \cite{Turney2010,hamilton2016diachronic,Dubossarsky2017}. The study of Diachronic Lexical Semantics \cite{Tahmasebi2018,Kutuzov2018}, which contributes towards detecting word-level language evolution, brings together researchers with broadly varying backgrounds from computational linguistics, cognitive science, statistics, mathematics, and historical linguistics, since the identification of words whose lexical semantics have changed over time has numerous downstream applications in various domains such as historical linguistics and NLP. Despite the increase in research interest, few tasks that track word meaning change over time have focused on non-English languages, while the comparison of different approaches in the same experimental and evaluation setting is still limited \cite{semeval}. The DIACR-Ita 2020 Task \cite{diacrita_evalita2020} aims to fill these gaps by focusing on the Italian language used during two different time periods and providing a single evaluation framework to researchers for testing their methods.


This work presents our approach towards detecting Italian words with altered lexical semantics during the two distinct time periods studied in the DIACR-Ita 2020 Shared Task. Our contribution focuses on evaluating findings
from previous studies, exploring  evaluation approaches for different methods and comparing their performance. We contrast several variants of training-testing words with different alignment approaches across two word embedding models, namely Skip-gram and Continuous Bag-of-Words \cite{mikolov2013distributed}. Our submission consisted of four models that showed the best average cosine similarity, calculated on the basis of their ability to accurately reconstruct the representations of Italian stop-words across the two  periods of time under study. Our best performing model uses a Continuous Bag-of-Words temporal compass model, adapted from the model introduced by \cite{Valerio2019Compass}. Our system ranked third in the task.

\section{Related Work}\label{sec:related-work}

Work related to unsupervised diachronic lexical semantics detection can be divided into different approaches depending on the type of word representations used in a diachronic model (e.g., based on graphs or probability distributions \cite{frermann2016bayesian,Azarbonyad2017}, temporal dimensions \cite{basile2018exploiting}, frequencies or co-occurrence matrices \cite{sagi2009semantic,cook2010automatically}, neural- or Transformer-based \cite{hamilton2016diachronic,Tredici,shoemark2019room,schlechtweg2019wind,giulianelli-etal-2020-analysing}, etc.). In our work, we focus on dense word representations \cite{mikolov2013distributed}, due to their high effectiveness that has been demonstrated in prior work.

Systems operating on representations such as those derived from Skip-gram or Continuous Bag-of-Words leverage in most cases deterministic approaches using mathematical matrix transformations \cite{hamilton2016diachronic,Azarbonyad2017,tsakalidis2019mining}, such as Orthogonal Procrustes \cite{schonemann1966generalized}, or machine learning models \cite{tsakalidis2020autoencoding}. The goal of these approaches is to learn a mapping between the word vectors that have been trained independently by leveraging textual information from two or more different periods of time. The common standard for measuring the level of diachronic semantic change of a word under this setting is to use a similarity measure (e.g., cosine distance) on the aligned space  -- i.e., after the mapping step is complete  \cite{Turney2010}. 

\cite{Dubossarsky2017}  argue that using cosine distance introduces bias in the system triggered by word frequency variations. \cite{Tan} only use the vectors of the top frequent terms to find the transformation matrix, and then they calculate the similarity for the remaining terms after applying the transformation to the source matrix. Incremental update \cite{kim2014,Tredici} used the intersection of words between datasets in each time frame by initializing the word embedding from the previous time slice to compare the word shift cross different years instead of using matrix transformation.  Temporal  Word  Embeddings with a Compass (TWEC) \cite{Valerio2019Compass} approach uses an approach of freezing selected vectors based on model's architecture, it learn a parallel embedding for all time periods from a base embedding frozen vectors. 

Our approaches, detailed in Section 4, follow and compare different methodologies from prior work based on (a) Orthogonal Procrustes alignment, (b) machine learning models and (c) aligned word embeddings across different time periods.





\section{Task Description}

The task was introduced by \cite{cignarella2020sardistance} and is defined as follows:

\vskip1.5truemm
\begin{tcolorbox}[colback=white!5!white,colframe=white!75!black]
Given two diachronic textual data, an unsupervised diachronic lexical semantics classifier should be able to find the optimal mapping to compare the diachronic textual data and classify a set of test words to one of two classes: 0 for stable words and 1 for words whose meaning has shifted.
\end{tcolorbox}
\vskip1.5truemm

We were provided with the two corpora in the Italian language, each from a different time period, and we developed several methods in order to classify a word in the given test set as ``semantically shifted'' or ``stable'' across the two time periods. The test set included 18 observed words -- 12 stable and 6 semantically shifted examples.

\section{Our Approach}
Here we outline our approaches for detecting words whose lexical semantics have changed.  


\subsection{Generating Word Vectors}
Word representations $W_i$ at the period $T_i$ were generated in two ways:

\vspace{.1cm}
\noindent\textit{(a) IND}: via Continuous Bag of Words (CBOW) and Skip-gram (SG) \cite{mikolov2013distributed} applied to each year independently;

\vspace{.12cm}
\noindent\textit{(b) CMPS}: via the Temporal Word Embeddings with a Compass (TWEC) approach \cite{Valerio2019Compass}, where a single model (CBOW or SG) is first trained over the merged corpus; then, SG (or CBOW) is applied on the representations of each year independently, by initialising and freezing the weights of the model based on the output of the first base model pass and learning only the contextual part of the representations for that year.

In both cases,  we used gensim with default settings.\footnote{\url{https://radimrehurek.com/gensim/}} Sentences were tokenised using the simple split function for flattened sentences provided by the organisers, without any further pre-processing. Although there are many approaches to generate word representations (e.g., using syntactic rules), we focused on 1-gram representations using CBOW and SG, without considering words lemmas and Part-of-Speech tags.

\subsection{Measuring Semantic Change}
We employ the cosine similarity for measuring the level of semantic change of a word. Given two word vectors $w^{T0}$, $w^{T1}$, semantic change between them is defined as follows:
\begin{equation}
\scriptsize
\cos(w^{T0}, w^{T1})=\frac{w^{T0} \cdot w^{T1}}{\|w^{T0}\|\|w^{T1}\|}=\frac{\sum_{i=1} w^{T0}_{i} w^{T1}_{i}}{\sqrt{\sum_{i=1} {w^{T0}_{i}}^{2}} \sqrt{\sum_{i=1} {w^{T1}_{i}}^{2}}}
\end{equation}
Though alternative methods have been introduced in the literature (e.g., neighboring by pivoting the top five similar words \cite{Azarbonyad2017}), we opted for the similarity metric which is most widely used in related work \cite{hamilton2016diachronic,shoemark2019room,tsakalidis2019mining}.

\subsection{Evaluation Sets} 
The challenge is expecting the lexical change detection to be done in an unsupervised fashion (i.e., no word labels have been provided). Thus, we considered stop words\footnote{\url{https://github.com/stopwords-iso/stopwords-it}} (\textit{SW}) and all of the other common words (\textit{CW}) in $T_0$ and $T_1$ as our training and evaluation sets interchangeably.

\subsection{Semantic Change Detection Methods}

We employed the following approaches for detecting words whose lexical semantics have changed:

\vspace{.12cm}
\noindent\textit{\underline{(a) Orthogonal Procrustes (OP)}}:
Due to the stochastic nature of CBOW/SG, the resulting word vectors $W_0$ and $W_1$ in \textit{IND} were not aligned. Orthogonal Procrustes \cite{hamilton2016diachronic} tackles this issue by aligning $W_1$ based on $W_0$. The level of semantic shift of a word is calculated by measuring the cosine similarity between the aligned vectors. For evaluation purposes, we measured the cosine similarity of the stop words between the two aligned matrices. Higher values indicate a better model (i.e., stop words retain their meaning over time).

\vspace{.12cm}
\noindent \textit{\underline{(b) Feed-Forward Neural Network (FFNN)}}: We trained a FFNN that leverages \textit{IND} to predict $W_1$ based on $W_0$. The level of semantic shift of a word in a test set is calculated by measuring the cosine similarity between the predicted $W^*_1$ and $W_1$. For evaluation purposes, we measure the cosine similarity between the actual and predicted representations of words in $T_1$. Higher values for stop-words indicate a better model.

\vspace{.12cm}
\noindent \textit{\underline{(c) Linear Regression (LR)}}: We employed an ordinary linear mapping with least square error objective function.\footnote{\url{https://scikit-learn.org/stable/}} The task and the evaluation setting was identical to FFNN.

\vspace{.12cm}
\noindent \textit{\underline{(d) Temporal Word Embeddings with a Compass}} \textit{\underline{(TWEC)}} \cite{Valerio2019Compass}: Working on the \textit{CMPS} vectors, the level of semantic shift of a word is calculated by measuring the cosine similarity between $T_0$ and $T_1$ directly. 


\paragraph{Notation} 
In the rest of this paper, we denote a model $M$ trained on CW (SW) as $M\_CW$ ($M\_SW$). For the case of $OP$, the training process involves learning an alignment based on a specific word set ($CW$ or $SW$). Note that this notation does not apply for $TWEC$, since the word vectors in the two time periods can be directly compared against each other -- thus the level of semantic change can be calculated directly (i.e., there is no need to learn any mapping between $W_0$ and $W_1$). Finally, we add a subscript $_{CBOW}$ or $_{SG}$ to our models, denoting the type of algorithm that was used for generating the respective embeddings that are fed to our model.

\paragraph{Model Selection} We select to apply the models on the test set providing high average cosine similarity with stop words.

\subsection{Word Classification}
As per the task guidelines \cite{cignarella2020sardistance}, words can fall into one of the two categories: \textbf{0}: the target word does not change meaning between $T_0$ and $T_1$ and \textbf{1}: the target word changes its meaning  between $T_0$ and $T_1$. For all of our submitted models, we considered all the words with cosine similarity below the mean as shifted words and labelled them with 1. We further investigate the model's ability to detect words laying two standard deviations below the mean ($\mu-2\sigma$), a.k.a variance. Interestingly, some of the models including LR and FFNN\_CW$_{C\-BOW}$ showed an increase in accuracy.


\section{Results}\label{sec:results}

\begin{table*}[ht] 
\centering
\begin{adjustbox}{max width=\textwidth}
\begin{tabular}{|l|l|l|l|l|l|l|l|l|l|l|l|l|l|}
\hline
\multicolumn{2}{|c|}{\textbf{IND} } & \multicolumn{6}{c|}{\textbf{SG} }  & \multicolumn{6}{c|}{\textbf{C-BOW} }  \\
\hline \hline
\multicolumn{2}{|c|}{\textbf{}} & \multicolumn{3}{c|}{\textbf{Accuracy} }  & \multicolumn{3}{c|}{\textbf{Ranking} }  & \multicolumn{3}{c|}{\textbf{Accuracy} }  & \multicolumn{3}{c|}{\textbf{Ranking} }  \\
     
\hline \hline
  train.   & \textbf{M} & $CS_{avg}^{SW}$ & \%$\mu$  & \%$\mu-2\sigma$ & \%$\mu_{rank}$ & $R_{p50}$  & $R_{\downarrow6}$ & $CS_{avg}^{SW}$  & \%$\mu$ & \%$\mu-2\sigma$ & \%$\mu_{rank}$ & $R_{p50}$ & $R_{\downarrow6}$    \\ \hline \hline
     
\textit{SW} & \textit{\textbf{OP}} &
  0.748 &
  \textbf{0.778} &
  \textbf{0.667} &
  \textbf{0.222} &
  \textbf{1.000} &
  \textbf{0.667} &
  0.784 &
  \textbf{0.778} &
  0.667 &
  \textbf{0.270} &
  \textbf{1.000} &
  \textbf{0.833} \\
 & \textbf{\textit{LR}} &
  \textbf{0.854} &
  0.333 &
  0.389 &
  0.373 &
  0.833 &
  0.500 &
  \textbf{0.795} &
  0.500 &
  \textbf{0.778} &
  \textbf{0.278} &
  0.833 &
  0.500 \\
   & \textbf{\textit{FFNN}} & \textbf{0.769} & 0.333 & 0.333          & 0.373          & 0.833 & 0.500 & 0.709 & 0.556 & \textbf{0.722} & 0.341 & 0.833 & 0.500 \\ \hline \hline
\textit{CW} & \textbf{\textit{OP}} & 0.464          & 0.389 & \textbf{0.778} & 0.381          & 0.667          & 0.500 & 0.289 & 0.611 & 0.667          & 0.397 & 0.833 & 0.333 \\
   & \textbf{\textit{LR}} & 0.409          & 0.333 & 0.444          & 0.508          & 0.500          & 0.333 & 0.146 & 0.333 & 0.444          & 0.381 & 0.667 & 0.667 \\
   & \textbf{\textit{FFNN}} & 0.658          & 0.333 & 0.389          & \textbf{0.317} & \textbf{1.000} & 0.500 & 0.621 & 0.333 & \textbf{0.722} & 0.317 & 0.833 & 0.500 \\ \hline \hline
 & \textbf{\textit{TWEC}} &
  0.722 &
  \textbf{0.722} &
  \textbf{0.667} &
  \textbf{0.317} &
  0.833 &
  \textbf{0.667} &
  \textbf{0.833} &
  \textbf{0.833$_*$} &
  0.667 &
  0.286 &
  \textbf{1.000} &
  \textbf{0.667} \\

\hline
\end{tabular}
\end{adjustbox}
\caption{Performance of our models using different evaluations methods. \textit{($_*$) best submission}.}
\label{tab:evaluation-table}
\end{table*}
\begin{figure}[htp]
\center
\subfloat{%
  \includegraphics[width=1\columnwidth]{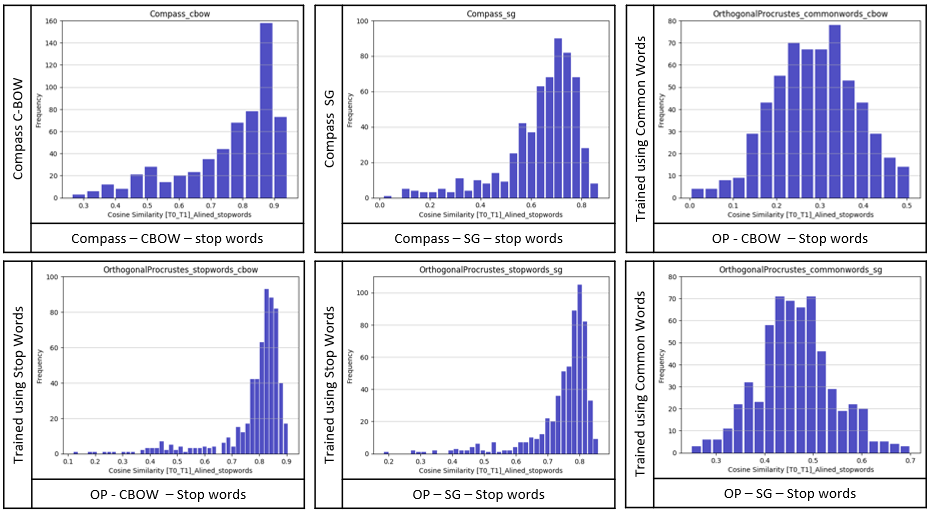}%
}
\label{fig:sp-similarity}
\caption{Frequency of stop words by their cosine similarity scores, where each subfigure pertains to a different model.}
\end{figure}

The results are shown in Table \ref{tab:evaluation-table}, where we split our results based on model \#M architecture, $_{SG}$ and $_{CBOW}$ and model's training word sets, Stop-Words (SW) and Common-Words (CW). For models based on linear transformation, our top performing models scored below average cosine similarity, TWEC$_{CBOW}$ ($0.833$), OP\_SW$_{SG}$($0.778$), OP\_SW$_{CBOW}$ ($0.778$), TWEC$_{SG}$($0.722$). As shown in Figure \ref{fig:sp-similarity}, we observe that these models tend to have skewed distributions for stop words, where the vast majority of stop words are assigned high cosine similarity scores. However, other models did not show this skewness, e.g. OP\_CW$_{SG}$($0.389$) and OP\_CW$_{CBOW}$($0.611$). When labeling the change based on variance ($\mu-2\sigma$), as in outlier detection, some models showed an increase from the dummy classifier's performance. For instance, OP\_CW$_{sg}$ showed an increase on performance from ($0.389$) to ($0.778$) showing that those with low average cosine similarity lay out in the tail from majority similarity. Similarly, models based on reducing the similarity error between the predicted and actual vectors, e.g. LR and FFNN considering the outlier detection methodology, tend to achieve better performance, including LR\_SW$_{CBOW}$, FFNN\_SW$_{CBOW}$ and FFNN\_CW$_{CBOW}$ where LR\_SW$_{CBOW}$ showed an increase from frequency classifier's baseline ($0.500$) to ($0.778$), and LR\_SW$_{CBOW}$ showed an increase from dummy classifier performance ($0.333$) to ($0.722$).


Ranking methods, average ranking ($\mu_{rank}$) and Recall ($R$), expect prior knowledge about the evaluation labels to make them useful for evaluating the reliability of the model of interest. For that, we further investigate the reliability of our experiment models, using $\mu\_{rank}$ and $R$ at $\%50$ (R$_{p50}$) and $\%30$ (R$_{\downarrow6}$). Although using (R$_{p50}$) signal OP\_SW$_{SG}$, OP\_SW$_{CBOW}$, FFNN\_CW$_{SG}$, TWEC$_{CBOW}$ as equalliy good, $\mu_{rank}$ ranked top models as OP\_SW$_{SG}$, OP\_SW$_{CBOW}$, LR\_SW$_{CBOW}$ then TWEC$_{CBOW}$ with (0.222, 0.270, 0.278 and 0.286), respectively. Additionally, under extreme conditions, OP\_SW$_{CBOW}$ ranked better than all including TWEC$_{CBOW}$. This shows that under extreme conditions, a good method is the one which keeps providing out of distribution signals to changing words and that needs to take a careful consideration about the distribution of the words before and after the alignments as in OP. In general, CBoW-based models showed better performance than SG-based models with average accuracy of (\%$\mu$ 0.564 and \%$\mu-2\sigma$ 0.667) compared to (\%$\mu$ 0.460 and $\mu-2\sigma$ 0.524) for words labelled by mean and variance, respectively. Further, alignment using non-changing words (e.g. \textit{stop-words}) yields higher performance than using all common words with average cosine similarity for stop words as ($CS_{avg}^{SW}$ 0.777) compared to ($CS_{avg}^{SW}$ 0.431), which is expected because SW-based models learns the optimal mapping with less noise than CW-based models.

\section{Discussion}\label{sec:discussion}
Our work provides a comprehensive analysis for Italian lexical diachronic methods introduced from previous work. For models that are based on matrix linear transformation including TWEC and OP, we find a relation between high average stop words similarity and accuracy. Further, C-BOW tends to achieve better results than the SG architecture for most experiments.  Visually, we find that a visibly skewed distribution showing the tendency of stop words to have high cosine similarity scores leads to effective means for capturing semantic shift. We also showed that by evaluating the models using different methods, TWEC$_{CBOW}$ achieved top performance. Followed by OP\_SW and OP\_CW$_{SG}$, and LR using outlier detection methodology. Further, FFNN showed high recall (R$_{p50}$) by ranking changed words with lowest cosine similarity on testing set similar to OP\_SW and TWEC$_{CBOW}$. This provides promising insights encouraging further investigation of neural network models using different languages and larger datasets. 




\section{Conclusions}
\label{sec:Conclusions}

In this report, we describe and compare our models submitted to the DIACR-Ita 2020 shared task, which assessed the ability to classify semantic-shift of words in Italian. We show that the TWEC model yields better performance than  Orthogonal Procrustes, labelling all words scored below average cosine similarity as semantically shifted words, i.e. words with altered semantics over the two time periods. Additionally, we showed that using an outlier detection methodology yields better results in prediction-based models such as Linear Regression and Feed-Forward Neural Network, boosting the performance significantly compared to the baselines and dummy classifier.

In the future we aim to focus on fine tuning SoTa pre-trained language models such as ELMo and BERT for word level semantics-shift detection as well as investigating the ability of dynamic graph models on capturing word evolution.

\section{Acknowledgments}
This research utilised Queen Mary's Apocrita HPC facility, supported by QMUL Research-IT.

\bibliographystyle{acl}
\bibliography{library}








\end{document}